\documentclass[10pt,journal,compsoc]{IEEEtran}

\usepackage{epsfig}
\usepackage{graphicx}
\usepackage{amsmath}
\usepackage{amssymb}
\usepackage{bm}
\usepackage{multirow}
\usepackage{multicol}
\usepackage{colortbl,hhline}
\usepackage{booktabs}
\usepackage{array,eucal}
\newcommand{\xie}[1]{{\color{black} #1}}

\renewcommand{\mathbb}[1]{\mathsf{#1}}
\def\REAL{{\mathcal R}}

\usepackage[margin=4pt,font=small,labelfont=bf,labelsep=endash,tableposition=top]{caption}
\usepackage[pagebackref=true,breaklinks=true,letterpaper=true,colorlinks,bookmarks=false]{hyperref}

\ifCLASSOPTIONcompsoc
  \usepackage[nocompress]{cite}
\else
  \usepackage{cite}
\fi

\begin{document}

\title{Learning from partially labeled data for multi-organ and tumor segmentation}

\author{Yutong~Xie, ~~
        Jianpeng~Zhang, ~~
        Yong~Xia, ~~
        Chunhua~Shen
\IEEEcompsocitemizethanks{\IEEEcompsocthanksitem 
Y. Xie was with
School of Computer Science and Engineering, Northwestern Polytechnical University, China. She is 
currently 
with
The University of Adelaide, Australia. (e-mail: yutong.xie678@gmail.com)
\IEEEcompsocthanksitem 
J. Zhang was with 
School of Computer Science and Engineering, Northwestern Polytechnical University, China. He is 
currently 
with the Alibaba DAMO Academy, China. (e-mail: jianpeng.zhang0@gmail.com)
\IEEEcompsocthanksitem 
Y. Xia is with National Engineering Laboratory for Integrated Aero-Space-Ground-Ocean Big Data Application Technology, School of Computer Science and Engineering, Northwestern Polytechnical University, China. (e-mail: yxia@nwpu.edu.cn)
\IEEEcompsocthanksitem 
C. Shen is with Zhejiang University, China. (e-mail: chunhua@me.com)
\IEEEcompsocthanksitem 
Corresponding authors: Y. Xia and C. Shen. Y. Xie and J. Zhang contributed equally.
Part of this work was done when J. Zhang and C. Shen were with The University of Adelaide. 
}
}

\markboth{ }%
{Zhang \MakeLowercase{\textit{et al.}}: 
Learning from partially labeled data for multi-organ and tumor segmentation
}

\IEEEtitleabstractindextext{%
\begin{abstract}
Medical image benchmarks for the segmentation of organs and tumors suffer from the partially labeling issue due to its intensive cost of labor and expertise. Current mainstream approaches follow the practice of one network solving one task. With this pipeline, not only the performance is limited by the typically small dataset of a single task, but also the computation cost linearly increases with the number of tasks. To address this, we propose a Transformer based dynamic on-demand network (TransDoDNet) that learns to segment organs and tumors on multiple partially labeled datasets. Specifically, TransDoDNet has a hybrid backbone that is composed of the convolutional neural network and Transformer. A dynamic head enables the network to accomplish multiple segmentation tasks flexibly. Unlike existing approaches that fix kernels after training, the kernels in the dynamic head are generated adaptively by the Transformer, which employs the self-attention mechanism to model long-range organ-wise dependencies and decodes the organ embedding that can represent each organ. We create a large-scale partially labeled Multi-Organ and Tumor Segmentation benchmark, termed MOTS, and demonstrate the superior performance of our TransDoDNet over other competitors on seven organ and tumor segmentation tasks. This study also provides a general 3D medical image segmentation model, which has been pre-trained on the large-scale MOTS benchmark and has demonstrated advanced performance over BYOL, the current predominant self-supervised learning method. 
Code will be available at \url{https://git.io/DoDNet}.
\end{abstract}

}

\maketitle
\IEEEdisplaynontitleabstractindextext

\IEEEpeerreviewmaketitle

\begin{figure*}[t]
\centering 
\includegraphics[width=0.97\linewidth]{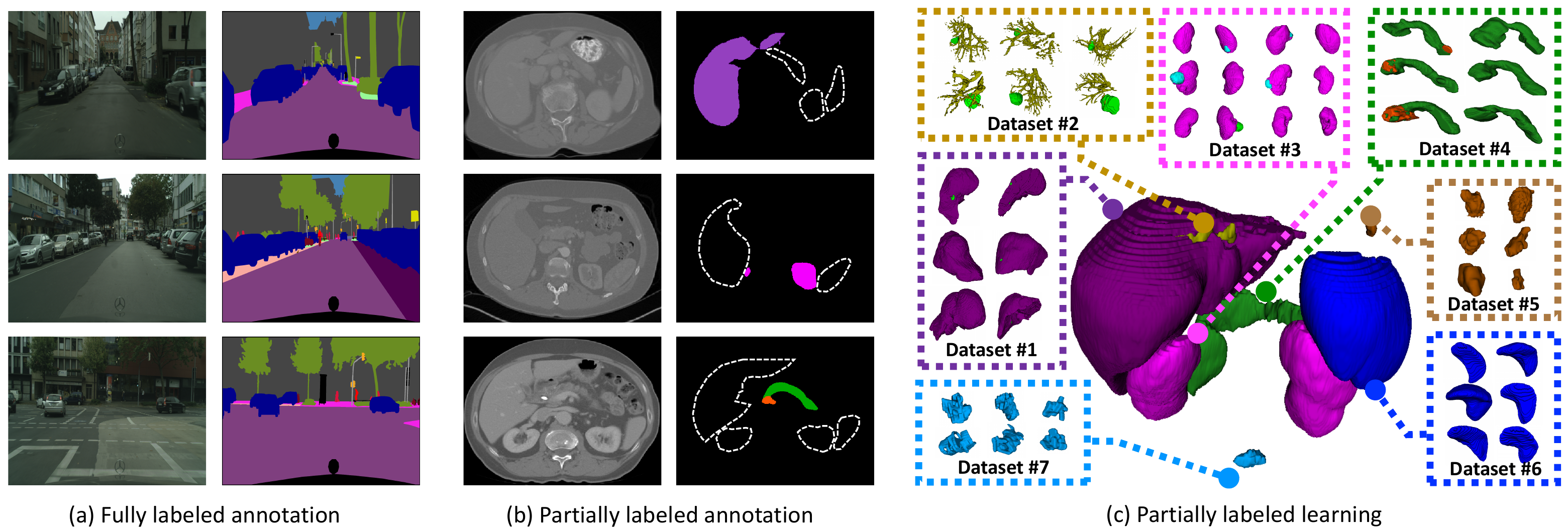}
\caption{
Comparison of fully labeled and partially labeled segmentation. In the fully labeled annotation, taking Cityscapes \cite{cordts2016cityscapes} for example (a), each pixel is assigned with a unique and true category label. As for the partially labeling, only the objects of interest have true category labels, other objects are regarded as the background. 
In this work, we aim to segment multiple organs and tumors by learning from several partially labeled datasets (c), each of which is originally specialized for the segmentation of a particular abdominal organ and$/$or related tumors. 
Here each color represents a partially labeled dataset. 
}
\label{fig:fig1}
\end{figure*}

\IEEEraisesectionheading{\section{Introduction}\label{sec:introduction}}

\IEEEPARstart{A}{utomated} segmentation of abdominal organs and tumors is one of the most fundamental yet challenging tasks in medical image analysis. It plays a pivotal role in a variety of computer-aided diagnosis jobs, including lesion contouring, surgical planning, and 3D reconstruction.
Most recently, it has witnessed rapid progress on this segmentation task, driven by both deep learning algorithms and public benchmark datasets \cite{pang2020ctumorgan,kavur2020chaos}. 
Constrained by the labor cost and expertise, 3D medical image segmentation benchmarks, however, severely suffer from the limited annotations on every single dataset. In this work, we attempt to collect multiple organs and tumor segmentation datasets to form a large-scale but more challenging multi-organ and tumor segmentation benchmark, termed MOTS. 

Unfortunately, most medical datasets were collected for the segmentation of only one type of organ and$/$or tumors, and all task-irrelevant objects were treated as the background (see Fig.~\ref{fig:fig1}). For instance, the LiTS dataset \cite{bilic2019liver} only has the annotations of liver and its tumors, and the KiTS dataset~\cite{heller2019kits19} only provides the annotations of kidneys and its tumors. As shown in Fig.~\ref{fig:fig1}, these partially labeled medical image datasets are different from fully labeled benchmarks in other computer vision areas, such as Cityscapes \cite{cordts2016cityscapes}, where multiple types of objects were annotated on each image. 
Therefore, one of the most significant challenges facing multi-organ and tumor segmentation is the so-called {\textit{partially labeling issue}}, \textit{i.e.}, learning the representation of multiple organs and tumors under the supervision of these partial annotations.

\begin{figure}[t]
\begin{center}
\includegraphics[width=1.0\linewidth]{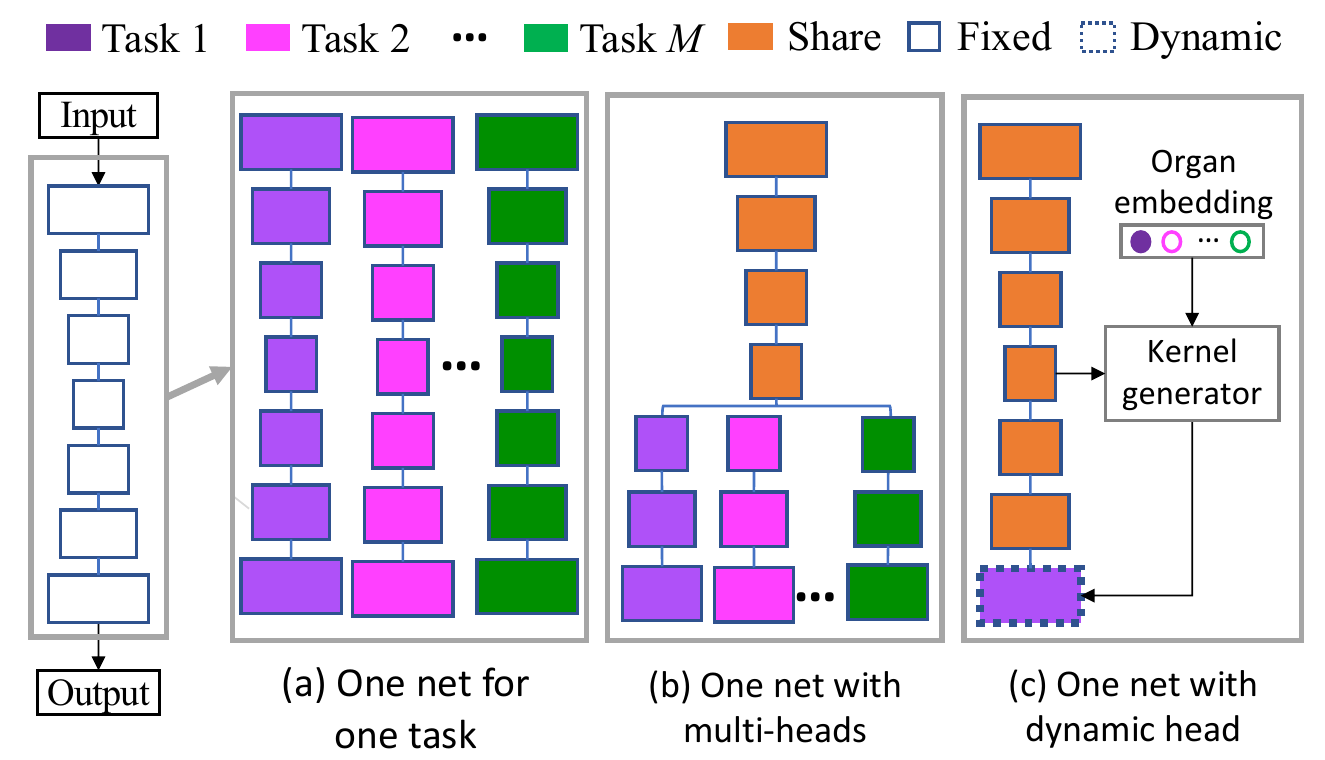}
\end{center}
\caption{Three types of methods to perform $M$ partially labeled segmentation tasks.
(a) One net for one task: training $M$ networks on $M$ partially labeled subsets, respectively;
(b) one net with multi-heads: training one network that consists of a shared encoder and $M$ task-specific decoders (heads), each head, together with the shared encoder, performing a partially labeled segmentation task; and
(c) the proposed one net with a dynamic head: it has an encoder, an organ embedding module, a kernel generator, and a dynamic segmentation head. The kernels in the dynamic head are conditioned on the input images and assigned organ embedding.}
\label{fig:dif2others}
\end{figure}

Mainstream approaches address this issue via separating the partially labeled dataset into several subsets and training a network on each subset for a specific segmentation task \cite{yu2019crossbar,isensee2021nnu,zhang2019light,myronenko20193d,zhu2019multi}, shown in Fig.~\ref{fig:dif2others}(a).
Following the one net for one task pipeline, each subset is fully labeled for the corresponding segmentation task. Such a strategy is intuitive but increases the computational complexity dramatically.
Another commonly-used solution is to design a multi-head network (see Fig.~\ref{fig:dif2others}(b)), which is composed of a shared encoder and multiple task-specific decoders (heads) \cite{chen2019med3d,fang2020multi,shi2020marginal}. In the training stage, feeding each partially labeled data to the network triggers the update of only one head, while other heads are frozen.
An obvious drawback of a multi-head network is that the number of segmentation heads increases with the number of tasks, leading to higher computational and spatial complexity. Besides, the inflexible multi-head architecture is not easy to be extended to new tasks.

\begin{figure}[t]
\begin{center}
\includegraphics[width=1.0\linewidth]{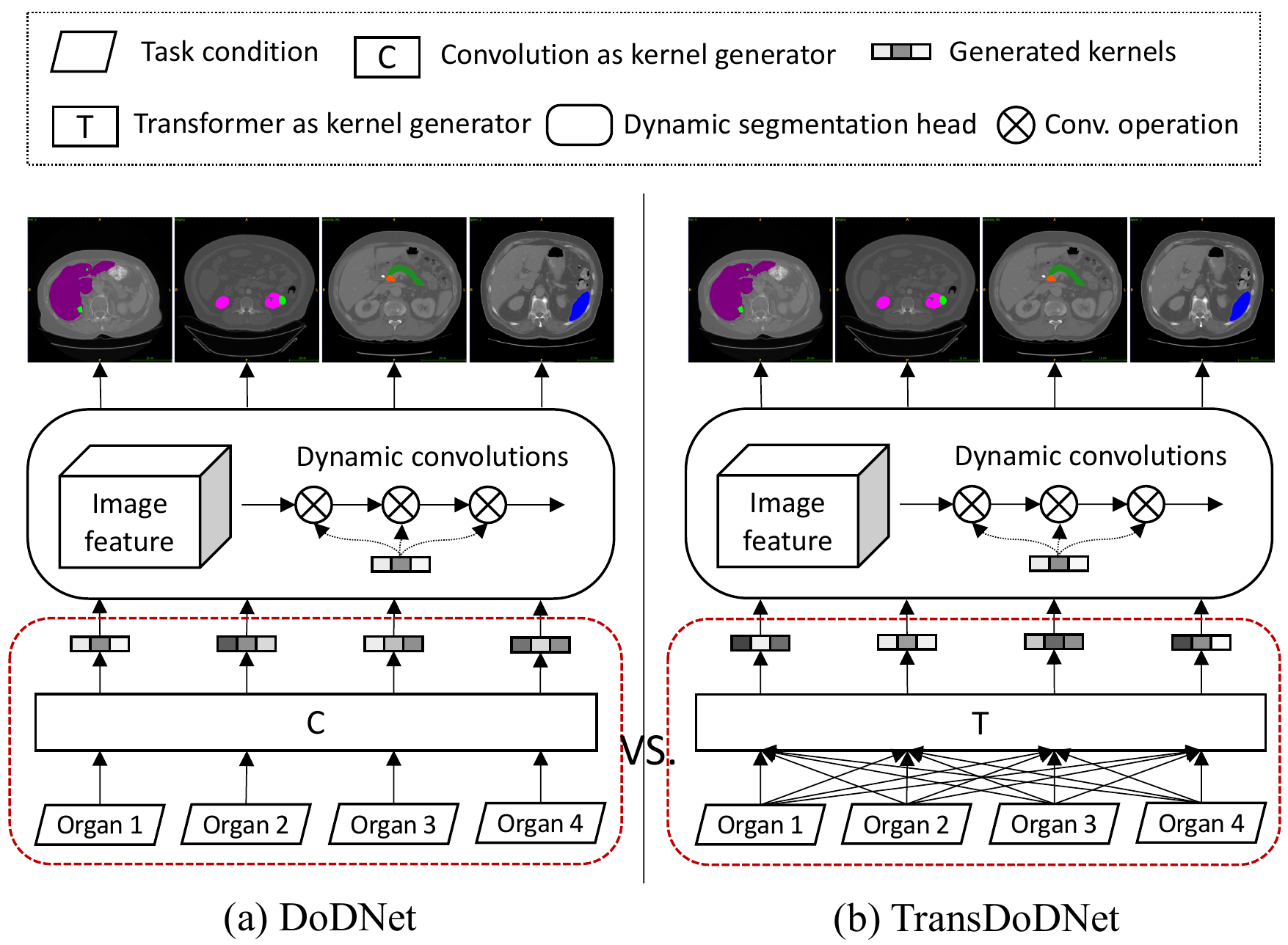}
\end{center}
\caption{Comparison of different kernel generation strategies. (a) DoDNet employs convolutions to generate specific kernels for each organ individually; and (b) TransDoDNet uses the Transformer as a kernel generator, enabling to model long-range organ-wise dependencies.}
\label{fig:controller}
\end{figure}

In our pilot study \cite{zhang2021dodnet}, we proposed a dynamic on-demand network (DoDNet), which can be trained on partially labeled datasets for multi-organ and tumor segmentation.
DoDNet has a simple but efficient encoder-decoder architecture. Different from previous solutions, its decoder is followed by a single but dynamic segmentation head (see Fig.~\ref{fig:dif2others}(c)).
We designed a convolution-based kernel generator to generate adaptively the kernels of the dynamic head for the segmentation of each organ, as shown in Fig.~\ref{fig:controller}(a).
Consequently, the dynamic segmentation head enables DoDNet to segment multiple organs and tumors as done by multiple networks or a multi-head network. 
However, these task-specific kernels are generated individually, ignoring the organ-wise dependencies. 

In this paper, we address this issue by introducing a novel Transformer based kernel generator, resulting in TransDoDNet.
With the self-attention mechanism, TransDoDNet is superior to its predecessor DoDNet in modeling the long-range organ-wise dependencies, as shown in Fig.~\ref{fig:controller}(b). 
The kernels generated by Transformer are not only conditioned on its own task but also mutually influenced by the dependencies on other tasks. 
We evaluate the effectiveness of TransDoDNet on MOTS, which is an ensemble of multiple organs and tumors segmentation benchmarks, involving the liver and tumors, kidneys and tumors, hepatic vessels and tumors, pancreas and tumors, colon tumors, and spleen. 
We also transfer the weights of TransDoDNet pre-trained on MOTS to downstream annotation-limited medical image segmentation tasks. The results show that our model has an excellent generalization ability, beating the current predominant self-supervised learning method \cite{BYOL}. 
Our contributions are four-fold.
\begin{itemize}
\item We create a large-scale multi-organ and tumor segmentation dataset by combining multiple partially labeled ones, resulting in a new and challenging medical image segmentation benchmark called MOTS. 
\item We address the partially labeling issue from a new perspective, \textit{i.e.}, proposing a single network that has a dynamic segmentation head to segment multiple organs and tumors as done by multiple networks or a multi-head network.
\item We are the first to employ the Transformer, which has a comprehensive perception of the global task-contextual information, as a kernel generator to achieve dynamic convolutions, contributing to better performance.
\item A byproduct of this work, \textit{i.e.}, the model weights pre-trained on MOTS, can be transferred to downstream 3D image segmentation tasks even with different image modalities, and hence is beneficial for the community.
\end{itemize}

\begin{figure*}[t]
\begin{center}
\includegraphics[width=1.0\linewidth]{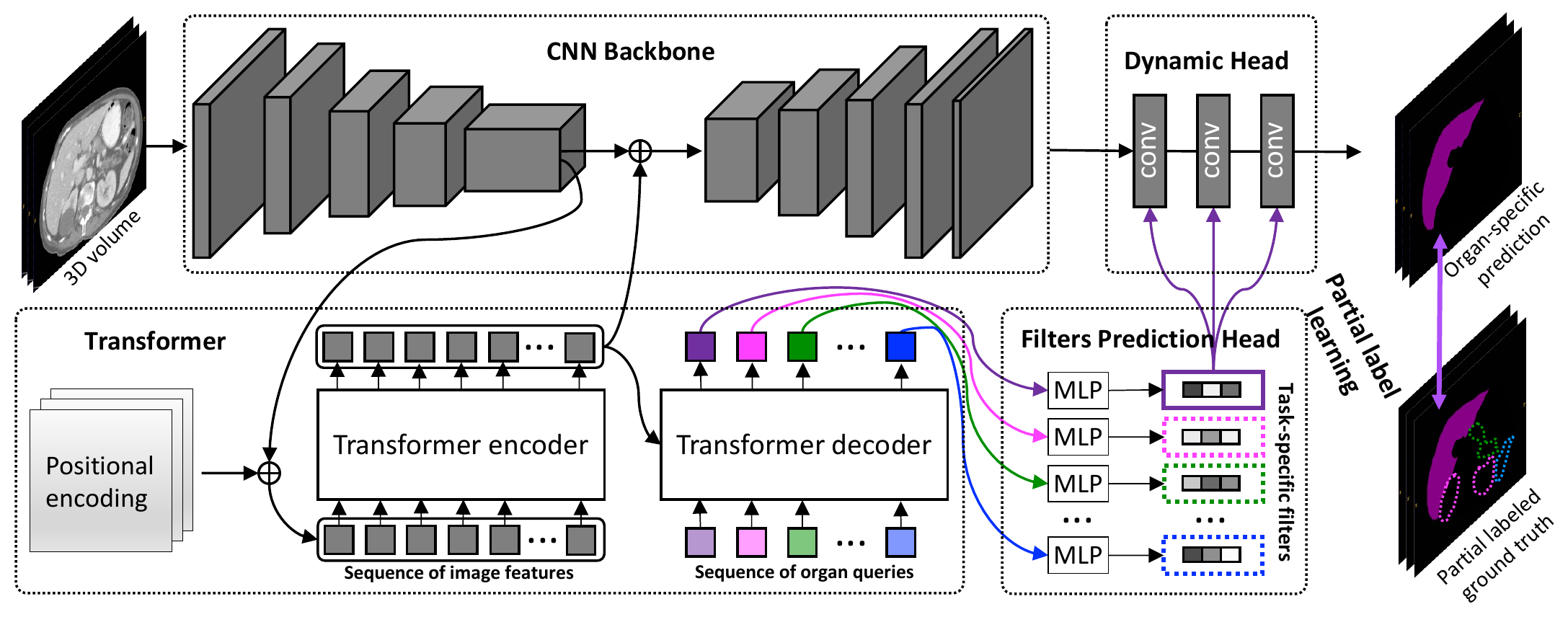}
\end{center}
\caption{Diagram of proposed TransDoDNet.}
\label{fig:framework}
\end{figure*}

\section{Related Work}
\subsection{Partially Labeled Medical Image Segmentation}
Segmentation of multiple organs and tumors is a generally recognized difficulty in medical image analysis \cite{Xie_multiorgan,zhang2020block,wang2019abdominal,schoppe2020deep}, particularly when there are no large-scale fully labeled datasets. Although several partially labeled datasets are available, each of them is specialized for the segmentation of one particular organ and$/$or tumors.
Accordingly, a segmentation model is usually trained on one partially labeled dataset, and hence is only able to segment one particular organ and tumors, such as the liver and liver tumors \cite{li2018h, zhang2019light, seo2019modified, tang2020net}, kidneys and kidney tumors \cite{myronenko20193d, Hou_kidney_isbi}. Training multiple networks, however, requires high computational resources and has poor scalability. 

To address this issue, many attempts have been made to explore multiple partially labeled datasets in a more efficient manner. 
Chen \textit{et al.} \cite{chen2019med3d} collected multiple partially labeled datasets from different medical domains, and co-trained a heterogeneous 3D network on them, which is specially designed with a task-shared encoder and task-specific decoders for eight segmentation tasks. 
Huang \textit{et al.} \cite{huang2020multi} proposed to co-train a pair of weight-averaged models for unified multi-organ segmentation on few-organ datasets. 
Zhou \textit{et al.} \cite{zhou2019prior} first approximated anatomical priors of the size of abdominal organs on a fully labeled dataset, and then regularized the organ size distributions on several partially labeled datasets. 
Fang and Yan \cite{fang2020multi} treated the voxels with unknown labels as the background, and then proposed the target adaptive loss (TAL) for a segmentation network that is trained on multiple partially labeled datasets. 
Shi \textit{et al.} \cite{shi2020marginal} merged unlabeled organs with the background and imposed an exclusive constraint on each voxel (\textit{i.e.} each voxel belongs to either one organ or the background) to learn a segmentation model jointly on a fully labeled dataset and several partially labeled datasets. 
To learn multi-class segmentation from single-class datasets, Dmitriev \textit{et al.} \cite{dmitriev2019learning} utilized the segmentation task as a prior and incorporated it into the intermediate activation signal.

The proposed TransDoDNet is different from these methods in four main aspects:
(1) In \cite{fang2020multi,shi2020marginal}, the partially labeling issue is formulated into a multi-class segmentation task, and unlabeled organs are treated as the background. This formulation may be misleading since the unlabeled organ in one dataset is indeed the foreground in another dataset. To amend this error, we formulate the partially labeling issue as a single-class segmentation task, aiming to segment each organ respectively;
(2) Most related work adopts a multi-head architecture, which is composed of a shared backbone network and multiple segmentation heads for different tasks. Each head is either a decoder \cite{chen2019med3d} or the last segmentation layer \cite{fang2020multi,shi2020marginal}. In contrast, the proposed TransDoDNet is a single-head network, in which the head is dynamic and can be generated adaptively;
(3) Our TransDoDNet uses the dynamic segmentation head to address the partially labeling issue, instead of embedding the task prior into the encoder and/or decoder; 
(4) Most existing methods focus on multi-organ segmentation, while our TransDoDNet is designed for the segmentation of both organs and tumors, which is more challenging. 

\subsection{Dynamic Filter Learning}
Dynamic filter learning has drawn considerable research attention in the computer vision community due to its adaptive nature \cite{jia2016dynamic,yang2019condconv,chen2020dynamic,Tian2020CondInst,he2019dynamic,HDFNet_ECCV2020}.
Jia \textit{et al.} \cite{jia2016dynamic} designed a dynamic filter network, in which the filters are generated dynamically conditioned on the input. This design is more flexible than traditional convolutional networks, where the learned filters are fixed during the inference.
Yang \textit{et al.} \cite{yang2019condconv} introduced the conditionally parameterized convolution, which learns specialized convolutional kernels for each input and effectively increases the size and capacity of a convolutional neural network (CNN).
Chen \textit{et al.} \cite{chen2020dynamic} presented another dynamic network, which dynamically generates attention weights for multiple parallel convolution kernels and assembles these kernels to strengthen the representation capability. 
Pang \textit{et al.} \cite{HDFNet_ECCV2020} integrated the features of RGB images and depth images to generate dynamic filters for better use of cross-modal fusion information in RGB-D salient object detection. 
Tian \textit{et al.} \cite{Tian2020CondInst} applied the dynamic convolution to instance segmentation, where the filters in the mask head are dynamically generated for each target instance. 
These methods successfully employ the dynamic filer learning toward certain ends,
such as 
increasing the network flexibility \cite{jia2016dynamic}, enhancing the representation capacity \cite{yang2019condconv, chen2020dynamic}, integrating cross-modal fusion information \cite{HDFNet_ECCV2020}, or abandoning the use of instance-wise ROIs \cite{Tian2020CondInst}. Comparing with these works,
our 
work here 
differs as follows.
1) we employ the dynamic filter learning to address the partially labeling issue for 3D medical image segmentation; and 2) the dynamic filters generated by Transformer, lead to better organ-wise dependency modeling. 

\subsection{Vision Transformers}
The transformer is a popular architecture commonly used in natural language processing~\cite{vaswani2017attention}, owing to its superior ability to model long-range dependencies. 
Recently, Transformer has also been widely used in the computer vision community as an alternative of CNN, and achieves the comparable or even the state-of-the-art performance on various tasks, including image recognition~\cite{ViT,DeiT}, semantic segmentation~\cite{SETR,VisTR}, object detection~\cite{DETR,deformableDETR}, and low-level image processing~\cite{chen2020pre}. 
Besides, many attempts have been made to extend the applications of Transformer to medical image scenarios, like TransUnet~\cite{chen2021transunet}, and CoTr~\cite{xie2021cotr}. 
Different from CNN, Transformer relies on the attention mechanism that models the long-range dependencies from a sequence-to-sequence perspective. 
Considering the inherent organ-wise dependencies in the multi-organ and tumor segmentation scenario, we employ the Transformer to generate dynamic kernels for the dynamic segmentation head, which captures the global organ-wise contextual information, and then generates the organ-specific filters in a parallel mode. 
To the best of our knowledge, this is the first attempt to employ the Transformer as a kernel generator in dynamic filter learning.

\section{Our Approach}
\label{Sec.Approach}

\subsection{Problem Definition}
Multiple organs and tumor segmentation aim to predict the pixel-level masks of each organ and corresponding tumors in the image. Given a medical image $\bm X$ and its pixel-level annotations $\bm Y$ sampled from the dataset $\mathfrak{D}$, this task can be easily optimized by minimizing a loss, usually, Dice loss or cross-entropy loss, as following
\begin{equation}
\min_{{\Theta }} \mathbb{E}_{({\bm X},{\bm Y}) \sim \mathfrak{D}} [ \mathcal{L} (f({\bm X}; {\Theta}), {\bm Y})]
\end{equation}
In this case, $\bm Y$ should be a full annotation that covers the pixel-level mask of multiple organs and tumors.

Let's consider $M$ partially labeled datasets $\{ \mathfrak{D}_{1},$ $ \mathfrak{D}_{2},$ $..., \mathfrak{D}_{M} \}$, which were collected from $M$ organ and tumor segmentation tasks:
\[
\verb'{P1:liver&tumor; P2:kidney&tumor; ...}'.
\]
Here, $\mathfrak{D}_{m}=\{ \bm{X}_{i}, \bm{Y}_{i} \}_{i=1}^{n_m}$ represents the $m$-th partially labeled dataset that contains $n_m$ labeled images. 
Given a image sampled from $\mathfrak{D}_{m}$, we denote it as $\widetilde{\bm{X}} \in \REAL^{D\times W \times H}$, where $W \times H$ is the spatial size of each slice and $D$ is number of slices. The corresponding segmentation ground truth is $\widetilde{\bm{Y}} \in \{0,1,2\}^{D\times W\times H}$, where the label of each voxel belongs to \verb'{0:background; 1:organ; 2:tumor}'. 
Straightforwardly, this partially labeled multi-organ and tumor segmentation problem can be solved by training $m$ segmentation networks $\{ f_1, f_2, ..., f_m \}$ on $m$ datasets, respectively, shown as follows
\begin{equation}
\left\{\begin{matrix}
&\min_{{\Theta }_1} \mathbb{E}_{(\widetilde{\bm X},\widetilde{\bm Y}) \sim \mathfrak{D}_{1}} [ \mathcal{L} (f_1(\widetilde{\bm X}; {\Theta}_1), \widetilde{\bm Y})] \\ 
&\vdots \\ 
&\min_{{\Theta }_m} \mathbb{E}_{(\widetilde{\bm X},\widetilde{\bm Y}) \sim \mathfrak{D}_{m}} [ \mathcal{L} (f_m(\widetilde{\bm X}; {\Theta}_m), \widetilde{\bm 
Y})] & 
\end{matrix}\right.
\end{equation}
where $\mathcal{L}$ represents the loss function of each network, $\{ \Theta_1, \Theta_2, ..., \Theta_m \}$ represent the parameters of these $m$ networks. 
In this work, we attempt to address this problem using only one single network $f$ with the parameter $\Theta$, which can be formally expressed as  
\begin{equation}
\min_{{\Theta }} \mathbb{E}_{\mathfrak{D}_{k}} [ \mathbb{E}_{(\widetilde{\bm X},\widetilde{\bm Y}) \sim \mathfrak{D}_{k}} [ \mathcal{L} (f(\widetilde{\bm X}; \Theta), \widetilde{\bm Y})]]
\end{equation}
The TransDoDNet proposed here for this purpose consists of a shared CNN encoder-decoder, a Transformer, a filters prediction head, and a dynamic segmentation head (see Fig.~\ref{fig:framework}). We now delve into the details of each part.

\subsection{Architecture}
\label{Sec.encoder-decoder}

The main component of TransDoDNet is a shared CNN encoder-decoder that has a U-like architecture \cite{ronneberger2015u}. 
The encoder is composed of repeated applications of 3D residual blocks \cite{he2016deep}, which has 50 learnable layers. 
Different from the vanilla ResNet, we employ the instance normalization \cite{instancenorm} to replace the batch normalization as done in \cite{isensee2021nnu}. Such a normalization strategy enables our model to be trained with a very small batch size. 
As for the input image $\bm{X}$, its feature map is calculated by
\begin{equation}
\bm{F} = f_E(\bm{X}; \bm{\theta}_{E})
\end{equation}
where $\bm{F} \in \REAL^{C_1\times D\times W \times H}$, and $\bm{\theta}_{E}$ represents the CNN encoder parameters. Besides, we employ the Transformer at the bottleneck to model the long-range feature dependencies, expressed by
\begin{equation}
\bm{Z} = f_{TE}(\bm{F}; \bm{\theta}_{TE})
\end{equation}
where $\bm{\theta}_{TE}$ represents the parameters of Transformer encoder. The details of the Transformer are presented in Sec.~\ref{sec.Transformer}. 

To decode the image feature map to the segmentation mask, we upsample the feature map to improve its resolution and halve its channel number step by step. At each step, the upsampled feature map is first summed with the corresponding low-level feature map from the encoder and then refined by a residual block. After the feed-forward of the decoder, we have
\begin{equation}
\bm{G} = f_D(\bm{F} + \bm{Z}; \bm{\theta}_{D})
\end{equation}
where $\bm{G} \in \REAL^{C_2\times D\times W \times H}$ is the pre-segmentation feature map, $\bm{\theta}_{D}$ represents all decoder parameters, and the channel number $C_2$ is set to 8 (see ablation study in Sec.~\ref{Sec.depth8}). 

The CNN encoder-decoder aims to generate $\bm{G}$, which is supposed to be rich-semantic and not subject to a specific task, \textit{i.e.}, containing the semantic information of multiple organs and tumors.

\subsection{Transformer based Kernel Generator}
\label{sec.Transformer}
\subsubsection{Volume positional encoding}
To feed the 3D feature map $\bm{F} $ into the Transformer, we 
first reduce the channel dimension from $C_1$ to $d$ by using a convolution layer, \textit{i.e.}, $\phi(\bm{F}): \bm{F} \in \REAL^{C_1\times D\times W \times H} \rightarrow \bm{S} \in \REAL^{d\times D\times W \times H}$, where $d \ll   C_1$.
Then we flatten the inter-slice and spatial dimensions and transpose it to a 1D sequence, $\bm{S} \in \REAL^{(D\times W \times H) \times d}$. 
However, such a 1D sequence is a collection of pixel-level feature maps, suffering a lack of spatial structure information. 

To address this issue, we augment the pixel-wise feature maps with the positional encoding that contains three dimensional positional information, including inter-slice, spatial width, and spatial height.  
We employ the fixed positional encoding strategy as done in \cite{vaswani2017attention}, and extend it to three dimensions to work on the 3D scenario. 
For each dimension $\# \in \{D, W, H\}$, its fixed positional encoding is calculated by
\begin{equation}
\left\{\begin{matrix}
&PE_{(pos, 2k)}^{\#} = \sin(pos / 10000^{2k/\frac{d}{3}}) \\ 
&PE_{(pos, 2k+1)}^{\#} = \cos(pos / 10000^{2k/\frac{d}{3}}) & 
\end{matrix}\right.
\end{equation}
We concatenate the positional encodings of three dimensions and then flatten them as a sequence
\begin{equation}
  \bm{E}_{pos} = {\rm Flat}(PE^{D}||PE^{W}||PE^{H})
\end{equation}
where $||$ represents the concatenation operation, $\rm Flat(\cdot)$ is the flatten operation, and $\bm{E}_{pos} \in \REAL^{(D\times W \times H) \times d}$.

\subsubsection{Attention mechanism}
\noindent\textbf{Self-attention: }
The self-attention mechanism is one of the key components in the Transformer~\cite{vaswani2017attention}. It models the similarity information of pixel-wise feature maps to all positions, formulated as
\begin{equation}
\begin{split}
  &\mathbb{SA} (q,k,v) = \rho^o(head_1|| head_2||...||head_h) \\
  &head_i = {\rm softmax}(\frac{\rho^q(q)\rho^k(k)^\top}{\sqrt{d_k}})\rho^v(v)
\end{split}
\end{equation}
where $\rho^q: \REAL^{d}\rightarrow \REAL^{d_k}$, 
$\rho^k: \REAL^{d}\rightarrow \REAL^{d_k}$,
$\rho^v: \REAL^{d}\rightarrow \REAL^{d_v}$,
$\rho^o: \REAL^{hd_v}\rightarrow \REAL^{d}$ are learnable feature projections, and $d_k=d_v=\frac{d}{h}$. 
The transformer with the self-attention mechanism would look over all possible locations in the feature map to explore the pixel-to-pixel level attention relationship, suffering from the slow convergence and high computational complexity. 

\noindent\textbf{Deformable attention: }
Inspired by \cite{dai2017deformable,deformableDETR}, we adopt the deformable multi-head attention in the Transformer to facilitate efficient training. 
The deformable multi-head attention enables the model to efficiently and flexibly attend to information from different representation subspaces at a few sparse positions, \textit{i.e.}, a small set of key sampling locations automatically selected by the model itself. 
Given the feature map sequence $\bm{S}$, the deformable multi-head attention of query $q$ to a set of key elements $K$ is calculated as
\begin{equation}
\begin{split}
  &\mathbb{DA}(q,\bm{S})=\rho^o(head_1|| head_2||...||head_h) \\
  &head_i = \sum_{k=1}^{K}A(q)_{k}\cdot \rho^s(\bm{S}) (p + \triangle_{p_{k}})
\end{split}
\end{equation}
where $A(q)_{k}$ is the learnable attention weights, $\rho^o$ and $\rho^s$ are learnable feature projections, $p$ is a 3D reference point, and $\triangle_{p_{k}}$ denotes the learnable sampling offsets.

\noindent\textbf{Multi-scale deformable attention: }
We denote $\bm{S}'=\{\bm{S}_1, \bm{S}_2, ..., \bm{S}_L\}$ as the multi-scale feature maps extracted from the different CNN encoder stages. Here $L$ is the number of multi-scale levels. We extend the single-scale attention mechanism to a multi-scale version, expressed as 
\begin{equation}
\begin{split}
  &\mathbb{MSDA}(q,\bm{S}')=\rho^o(head_1|| head_2||...||head_h) \\
  &head_i = \sum_{l=1}^{L}\sum_{k=1}^{K}A(q)_{lk}\cdot \rho^s(\bm{S}'_l) (\sigma_l(p) + \triangle_{p_{lk}})
\end{split}
\end{equation}
where $\sigma_l$ re-scales $p$ to the $l$-th level feature. 
Compared to single-scale attention, multi-scale deformable attention is more flexible and powerful to observe all possible locations in the multi-scale feature maps. 

\subsubsection{Transformer encoder}
The Transformer encoder is composed of the alternating layers of multi-scale deformable attention layer and fully connected feed-forward network (FFN). 
To speed up the training process, residual connections \cite{he2016deep,vaswani2017attention} are performed across each of the multi-head attention layer and MLP layer, followed by the layer normalization ($\mathbb{N}$)~\cite{ba2016layer}. Specifically, we follow the deep norm strategy~\cite{wang2022deepnet} to stabilize the deep Transformers. 
Given a $\mathbb{I}_{enc}$-layer encoder, the feed-forward process of each Transformer layer can be formulated as 
\begin{equation}
\begin{split}
  & z_0 = \bm{S}' \\
  & q_i = z_{i-1} + \bm{E}_{pos}, x_i = z_{i-1} \\
  & z'_i = \mathbb{N}(\mathbb{MSDA}(q_i, x_i) + z_{i-1}*\alpha) \\
  & z_i = \mathbb{N}(\mathbb{FFN}(z'_i) + z'_i*\alpha)
\end{split}
\end{equation}
where $i=1,...,\mathbb{I}_{enc}$, and $\alpha$ is a scaling factor of residual connections that is conditioned on the number of Transformer encoder and decoder layers. 

\subsubsection{Transformer decoder}
The Transformer decoder follows a similar architecture as the encoder, including two multi-head attention layers and one MLP layer. 
Besides, we introduce the organ embeddings as the input of the decoder, denoted as $\bm{E}_q \in \REAL^{M\times d}$. Here $M$ is the total number of queries, which is fixed during the training. 
The organ embeddings are learned to decode features for each organ segmentation task. 
The Transformer decoder receives the organ queries and memory from the output of the Transformer encoder and produces a sequence of output embeddings, each of which can represent one organ in the input volumes. 
The calculation of the Transformer decoder can be formulated as 
\begin{equation}
\begin{split}
&t_0 = \varrho  \\
&v_i=t_{i-1}, q_i=k_i=t_{i-1} + \bm{E}_q \\
&t'_i=\mathbb{N}(\mathbb{SA}(q_i,k_i,v_i)+t_{i-1}*\alpha) \\
&q'_i=t'_i + \bm{E}_q, x'_l=z_{\mathbb{I}_{enc}} \\
&t_i = \mathbb{N}(\mathbb{MSDA}(q'_i,x'_i) + t'_i*\alpha) \\
&t_i = \mathbb{N}(\mathbb{FFN}(t_i)+t_i*\alpha)
\end{split}
\end{equation}
where $i=1,...,\mathbb{I}_{dec}$, and $\varrho $ is the random initialization of organ embeddings.

\subsection{Filters Prediction Head}
Transformer encodes the 3D feature maps and then decodes the organ embeddings into a sequence of output embeddings, \textit{i.e.}, $t_L \in \REAL^{M\times d}$, each of which can represent a specific organ. 
A single MLP is assigned to decode them into the task-specific filters for the dynamic segmentation head. 
\begin{equation}
\bm{\omega} = \mathrm{MLP}(t_{\mathbb{I}_{dec}}) \in \REAL^{N\times d_{F}}
\label{eq:filters}
\end{equation}
where $d_{F}$ is the dimension of generated filters.


\subsection{Dynamic Head}
\label{Sec.DynamaicHead}
During the partially supervised training, it is worthless to predict the organs and tumors whose annotations are not available. 
Therefore, a lightweight dynamic head is designed to enable specific kernels to be assigned to each task for the segmentation of a specific organ and tumors. The dynamic head contains three stacked convolutional layers with $1\times1\times1$ kernels. 
The kernel set in $h_d$ layers, denoted by $\bm{\omega} = \{\bm{\omega}_{1}, \bm{\omega}_{2}, ..., \bm{\omega}_{h_d} \}$, are dynamically generated by the filters prediction head according to the organ embeddings (see Eq.~\ref{eq:filters}).
The dynamic width is set to $h_w$ channels in front of $h_d-1$ layers, while the last layer is fixed to 2 channels, \textit{i.e.}, one channel for organ pixel-wise prediction and the other for tumor pixel-wise prediction. 
The total number of dynamic parameters is computed by $d_{F} = (h_w\times h_w+h_w)\times (h_d-1)+ (h_w\times 2+2)$. 
The prediction mask of the image $\bm{X}$ sampled from the partially labeled dataset $\mathfrak{D}_m$ is denoted as $\bm{P} \in  \REAL^{M\times 2\times D\times W \times H}$, where
\begin{equation}
  \bm{P}_i = \begin{cases}
    ((\bm{G} \ast \bm{\omega}_{1i}) \ast \bm{\omega}_{2i}) \ast \bm{\omega}_{3i} & \text{ if } i= m \\ 
    \; \; \; \; \; \; \; \; \; \; \; \; \; \;   \varnothing  & \text{ otherwise } 
    \end{cases}\end{equation}
Here, $i=1,2,...,M$, and $\ast$ represents the convolution.
Although each image requires a group of specific kernels for each task, the computation and memory cost of our lightweight dynamic head is negligible compared to the encoder-decoder (see Sec.~\ref{Sec.SpeedvsAccuracy}).

\subsection{Training and Testing}

For simplicity, we treat the segmentation of an organ and related tumors as two binary segmentation tasks, and jointly use the Dice loss and binary cross-entropy loss as the objective for each task. Given the predictions $\bm{P}$ and partial labeled ground truth $\bm{Y} \in  \REAL^{2\times D\times W \times H}$ sampled from $\mathfrak{D}_m$, the loss function is formulated as 
\begin{equation}
\begin{aligned}
\small
\mathcal{L}(\bm{P},\bm{Y}) = - &\sum_{k=1}^{2}[\frac{2  \sum_{p_i\in \bm{P}_{mk}, y_i \in \bm{Y}_k} p_i y_i}{ \sum_{p_i \in \bm{P}_{mk}, y_i \in \bm{Y}_{k}} (p_i + y_i + \epsilon)} \\
+ \sum_{p_i\in \bm{P}_{mk}, y_i \in \bm{Y}_{k}} & {(y_i \log p_i + (1-y_i)\log(1-p_i))}]
\end{aligned}
\end{equation}
where $\epsilon$ is a smoothing factor for Dice calculation. 
A simple strategy is used to process the tasks that only provide organ or tumor annotations, \textit{i.e.}, ignoring the predictions corresponding to unlabeled targets.
Taking colon tumor segmentation for example, the result of organ prediction, \textit{i.e.}, when $k=1$, is ignored during the loss computation and error back-propagation since the annotations of organs are unavailable. 

During inference, the proposed TransDoDNet is flexible to $M$ segmentation tasks. 
Given a test image, the pre-segmentation feature $\bm{G}$ is firstly extracted from the encoder-decoder network. 
Then, the Transformer generates the organ embeddings for the filter prediction head to generate segmentation kernels. 
Assigned with a task, its specific kernels are extracted and input to the dynamic segmentation head for the specific segmentation purpose. 
In addition, if $M$ tasks are all required, our TransDoDNet is able to simultaneously infer the dynamic head by using different kernels in a parallel manner. 
Compared to the CNN encoder-decoder, the dynamic head is so light that the repeated inference cost of dynamic heads is almost negligible.


\subsection{Transferring to Downstream Tasks}
\begin{figure}[h]
  \begin{center}
  \includegraphics[width=0.95\linewidth]{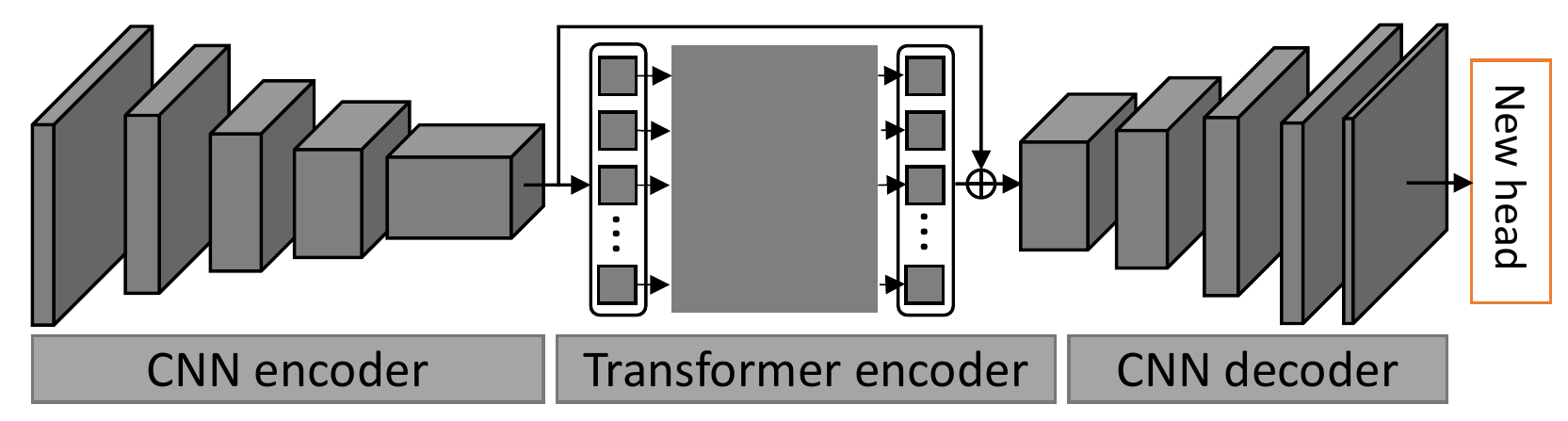}
  \end{center}
  \caption{Model architecture for downstream tasks.}
  \label{fig:downstream_archi}
  \end{figure}

TransDoDNet is proposed to address more challenging medical image segmentation tasks by using large-scale partially labeled datasets. Such a large-scale dataset is able to promote the model's generalization ability, which is also beneficial to the downstream tasks with limited annotations. 
To transfer the weights pre-trained on MOTS and fit them for the downstream tasks, we reuse the CNN encoder, Transformer encoder, and CNN decoder, while abandoning all other components in TransDoDNet. The new model architecture is shown in Fig.~\ref{fig:downstream_archi}. To adapt to a new segmentation task, a new classification head is added at the end of the CNN decoder. Note that the gray modules are well pre-trained on the large-scale partial labeled datasets.

\begin{table}
%
%
\small 
\caption{Details about MOTS dataset, including partial labels, available annotations, and number of training and test images. \checkmark means the annotations are available and $\times$ is the opposite.}
\begin{center}
\begin{tabular}{l|c|c|c|c}
\hline
\multirow{2}{*}{Partial-label task} & \multicolumn{2}{c|}{Annotations} & \multicolumn{2}{c}{\# Images} \\ \cline{2-5} 
 & Organ & Tumor & Training & Test \\ \hline
\#1 Liver & \checkmark & \checkmark & 104 & 27 \\ \hline
\#2 Kidney & \checkmark & \checkmark & 168 & 42 \\ \hline
\#3 Hepatic Vessel & \checkmark & \checkmark & 242 & 61 \\ \hline
\#4 Pancreas & \checkmark & \checkmark & 224 & 57 \\ \hline
\#5 Colon & $\times$ & \checkmark & 100 & 26 \\ \hline
\#6 Lung & $\times$ & \checkmark & 50 & 13 \\ \hline
\#7 Spleen & \checkmark & $\times$ & 32 & 9 \\ \hline
Total & - & - & 920 & 235 \\ \hline
\end{tabular}
\end{center}
\label{tab:MOTS_details}
\end{table}

\begin{table}[h] 
  \small 
  \caption{\xie{Dynamic head with different depths (\#layers), varying from 2 to 4. Here the head width is fixed to 8.}}
  \begin{center}
    \begin{tabular}{c|c|c|c}
      \hline
      Depth & Dyn. params & mDice & mHD \\ \hline
      2     & 90 & 72.11 & 21.46 \\ \hline 
      3     & 162 & \textbf{72.30} & \textbf{20.74} \\ \hline 
      4     & 234 & 71.43 & 22.54 \\ \hline 
      \end{tabular} 
  \end{center}
  \label{tab:ablation_depth}
  \end{table}

\begin{table}[h] 
    \small 
    \caption{Dynamic head with different widths (\#channels), varying from 4 to 8. Here the head depth is fixed to 3.}
    \begin{center}
      \begin{tabular}{c|c|c|c}
        \hline
        Width & Dyn. params & mDice & mHD \\ \hline
        4     & 50 & 71.51 & 25.29  \\ \hline 
        8     & 162 & \textbf{72.30} & \textbf{20.74} \\ \hline 
        16    & 578 & 71.97 & 23.57 \\ \hline 
        \end{tabular}
    \end{center}
    \label{tab:ablation_width}
    \end{table}

\section{Experiment}
\subsection{Experiment Setup}
\noindent\textbf{Dataset}: 
We build a large-scale partially labeled \textbf{M}ulti-\textbf{O}rgan and \textbf{T}umor \textbf{S}egmentation (MOTS) benchmark using multiple medical image segmentation datasets, including LiTS\cite{bilic2019liver}, KiTS \cite{heller2019kits19}, and Medical Segmentation Decathlon (MSD) \cite{simpson2019large}. 
MOTS is composed of seven partially labeled sub-datasets, involving seven organ and tumor segmentation tasks. 
There are 1155 3D abdominal CT scans collected from various clinical sites around the world, including 920 scans for training and 235 scans for test. 
More details are given in Table~\ref{tab:MOTS_details}. 
Each scan is re-sliced to the same voxel size of $1.5\times0.8\times0.8mm^3$.

Two public datasets are also used for the downstream tasks, including BCV \cite{BCV_benchmark} and BraTS \cite{brats_benchmark}. 
BCV was collected by the MICCAI 2015 Multi Atlas Labeling \textbf{B}eyond the \textbf{C}ranial \textbf{V}ault Challenge. BCV is composed of 50 abdominal CT scans, including 30 scans for training and 20 scans for test. Each training scan is paired with voxel-wise annotations of 13 organs, including the liver (Li), spleen (Sp), pancreas (Pa), right kidney (RK), left kidney (LK), gallbladder (Ga), esophagus (Es), stomach (St), aorta (Ao), inferior vena cava (IVC), portal vein and splenic vein (PSV), right adrenal gland (RAG), and left adrenal gland (LAG). 
BraTS was provided by the 2018 \textbf{B}rain \textbf{T}umor \textbf{S}egmentation Challenge. The aim of this challenge is to develop automated segmentation algorithms to delineate intrinsically heterogeneous brain tumors, \textit{i.e.}, (1) enhancing tumor (ET), (2) tumor core (TC) that consists of ET, necrotic and non-enhancing tumor core, and (3) whole tumor (WT) that contains TC and the peritumoral edema. This dataset provides the different imaging modalities, \textit{i.e.}, magnetic resonance imaging (MRI). Each case contains four MRI sequences, including the T1, T1c, T2, and FLAIR. All sequences were registered to the same anatomical template and interpolated to the same dimension of $155\times240\times240$ voxels and the same voxel size of $1.0\times1.0\times1.0$ $mm^3$. BraTS is composed of 285 training cases, including 210 cases of high-grade gliomas and 75 cases of low-grade gliomas, and 66 cases for online evaluation.
We evaluate the pre-trained weights of MOTS on these downstream segmentation tasks. 

\noindent\textbf{Evaluation metric}: 
The Dice similarity coefficient (Dice) and Hausdorff distance (HD) are used as performance metrics for this study. Dice measures the overlapping between a segmentation prediction and ground truth, and HD evaluates the quality of segmentation boundaries by computing the maximum distance between the predicted boundaries and ground truth. We assess all segmentation methods based on the mean Dice (mDice)) and mean HD (mHD) across all organ and tumor categories.

\noindent\textbf{Implementation details}: 
To filter irrelevant regions and simplify subsequent processing, we truncate the HU values in each scan to the range of $[-325, +325]$ and linearly normalize them to $[-1, +1]$.
Owing to the benefits of instance normalization \cite{instancenorm}, our model adopts the micro-batch training strategy with a small batch size of 2. 
The AdamW algorithm \cite{adamw} is adopted as the optimizer. The learning rate is initialized to $2\times10^{-4}$ and decayed according to a polynomial policy ${\rm lr}={\rm lr}_{init}\times(1-k/K)^{0.9}$, where the maximum epoch $K$ is set to 1,000. 
In the training stage, we randomly extract sub-volumes with the size of $64\times192\times192$ from CT scans as the input. 
To avoid the overfitting problem, we use the batchgenerator library \cite{isensee2020batchgenerators} to perform a wide variety of data augmentation, including randomly scaling, flipping, rotation, Gaussian noise, Gaussian blur, contrast, and brightness. 
In the test stage, we employ the sliding window based strategy and let the window size be $64\times256\times256$. 
To ensure a fair comparison, the same training strategies, including the data augmentation, learning rate, optimizer, and other settings, are applied to all competing models. 

\subsection{Ablation Study}
We split 20\% of training cases as validation data to perform the ablation study. We report the mean Dice and HD of 11 organs and tumors (listed in Table~\ref{tab:MOTS_details}) as two evaluation indicators. We set the maximum epoch to 500 for the efficient and fair ablation study. 

\subsubsection{Dynamic head}
We first investigate the effectiveness of the dynamic head, including the depth and width. Specifically, we keep the depth of the Transformer encoder and decoder to be 1, and the multi-scale level to be 1, \textit{i.e.}, single scale. We report the segmentation performance and total dynamic parameters of each setting for a clear comparison.

\noindent \textbf{Head depth}:  
\label{Sec.depth8}
In Table~\ref{tab:ablation_depth}, we compare the performance of the dynamic head with different depths, varying from 2 to 4. The width is fixed to 8 channels, except for the last layer, which has 2 channels. 
It shows that the best performance, including both mDice and mHD, is achieved when the head depth equals 3. Besides, the performance fluctuation is very small when the depth increases from 2 to 4. The results indicate the robustness of the dynamic head to the varying depth, which is empirically set to 3 for this study.

\noindent \textbf{Head width}: 
In Table~\ref{tab:ablation_width}, we compare the performance of the dynamic head with different widths, varying from 4 to 16. Here the depth is fixed to 3. 
It shows that the performance improves substantially when increasing the width from 4 to 8, but drops slightly when further increasing the width from 8 to 16. It suggests that the performance tends to become stable when the width of the dynamic head falls within a reasonable range. Considering both performance and complexity, we empirically set the width of the dynamic head to 8.

\subsubsection{Kernel generator}
In this section, we first compare the effectiveness of two different kernel generators, \textit{i.e.}, convolution and Transformer, and then investigate the key components of the Transformer based kernel generator, including Transformer depth, width, and multi-scale attention mechanism. 
Note that the depth and width of the dynamic head are uniformly set to 3-layer and 8-channel respectively in the rest of the experiments. 

\noindent \textbf{CNN vs. Transformer based kernel generator}: 
We compare the Transformer based kernel generator with the convolution based solution that was presented in our conference version. The convolution based method generates kernels conditioned on the independent task encoding, ignoring organ-wise dependencies. Differently, the Transformer based kernel generator is superior to modeling the global organ-wise contextual information, and hence segments the multi-organs and tumors more accurately. As shown in Table~\ref{tab:conv_vs_trans_kernel_generator}, both methods utilize the one-layer architecture in the kernel generator. The proposed TransDoDNet is assigned with a basic setting that has one Transformer encoder-decoder layer in the kernel generator, and the feature channel is set to 384. 
For a fair comparison, we scale up the convolution based DoDNet by increasing the convolution layers in the last backbone stage to achieve comparable parameters with TransDoDNet. 
We can see that the Transformer generator outperforms the convolution generator by a clear margin in all metrics (Transformer 72.30 mDice vs. convolution 70.98 mDice) with comparable parameters (Transformer 40.52M vs. convolution 42.49M). 

\begin{table}[]
  \caption{Comparison of the different kernel generators. Convolution* refers to the deepening DoDNet that has the comparable parameters with TransDoDNet. }
  \centering
  \begin{tabular}{c|c|c|c}
  \hline
  Kernel generator & Params & mDice & mHD   \\ \hline
  Convolution      & 35.31  & 70.52 & 26.40  \\ \hline
  Convolution*      & 42.49  & 70.98 & 23.95  \\ \hline
  Transformer      & 40.52 & \textbf{72.30} & \textbf{20.74} \\ \hline
  \end{tabular}
  \label{tab:conv_vs_trans_kernel_generator}
  \end{table}

\noindent \textbf{Transformer depth}: 
We compare the Transformer with different layers of encoder and decoder in Table~\ref{tab:TR_depth}. 
We observe better segmentation performance when increasing the Transformer encoder/decoder layers. However, this trend is weakened when the layer goes up to 6, which suffers from a little performance saturation. As the Transformer layer increases, the model is mired in the fast-growing parameters. 
We thus empirically set the layers of the Transformer encoder and decoder to be 3, balancing the model accuracy and complexity. 

\begin{table}[h] 
  \small 
  \caption{Transformer-based kernel generators with different encoder and decoder layers.}
  \begin{center}
    \begin{tabular}{c|c|c|c|c}
      \hline
      Encoder & Decoder & Params & mDice & mHD \\ \hline
      1       & 1       & \textbf{40.52} &  72.30   & 20.74  \\ \hline
      3       & 1       & 43.77 &  72.62   & 20.31  \\ \hline
      1       & 3       & 44.96 &  72.57   & 20.59  \\ \hline
      3       & 3       & 48.21 &  \textbf{73.37}   & \textbf{18.61}  \\ \hline
      6       & 6       & 59.74 &  73.07   & 19.31  \\ \hline
      \end{tabular}
  \end{center}
  \label{tab:TR_depth}
  \end{table}

\noindent \textbf{Transformer width}: 
Transformer width, \textit{i.e.}, feature channel, is also an important factor that determines the scale of Transformers. In Table~\ref{tab:TR_width}, we increase progressively the Transformer width from 96-channel to 768-channel, leading to a dramatic increase of parameters from 36.49 to 82.67 million. We observe that the performance deteriorates as the increase of width, and obtain the best performance when the width is 192. The massively increased parameters may bring a great difficulty to model optimization, thus failing to observe a performance improvement.

\begin{table}[h] 
  \small 
  \caption{Transformer with different width (channels).}
  \begin{center}
    \begin{tabular}{c|c|c|c}
      \hline
      Width & Params & mDice & mHD \\ \hline
      96 & \textbf{36.49} & 71.55 & 22.58 \\ \hline
      192 & 39.05 &  \textbf{73.79} &  18.95 \\ \hline
      384 & 48.21 &  73.37 & \textbf{18.61} \\ \hline
      768 & 82.67 &  72.02 & 23.18 \\ \hline
      \end{tabular}
  \end{center}
  \label{tab:TR_width}
  \end{table}

\noindent \textbf{Multi-scale attention vs. single-scale attention}: 
The proposed TransDoDNet is able to model the long-range dependencies across multiple scales, which is beneficial for segmenting organs and tumors with diverse object scales. 
Compared to the single-scale attention, as shown in Table~\ref{tab:MS_levels}, the multi-scale attention with 3 levels improves the mDice from 72.80\% to 73.79\% and reduces the mHD from 20.27 to 18.95. 

\begin{table}[h]
\small 
\caption{TransDoDNet with different multi-scale levels.}
\begin{center}
  \begin{tabular}{c|c|c}
    \hline
    Multi-scale levels & mDice & mHD \\ \hline
    1                     & 72.80     & 20.27      \\ \hline
    2                     & 73.36     & 19.11      \\ \hline
    3                     & \textbf{73.79}    & \textbf{18.95}  \\ \hline
    \end{tabular}
\end{center}
\label{tab:MS_levels}
\end{table}

\subsubsection{Comparison of three variants of TransDoDNet}
\begin{figure}[h]
\begin{center}
\includegraphics[width=1.0\linewidth]{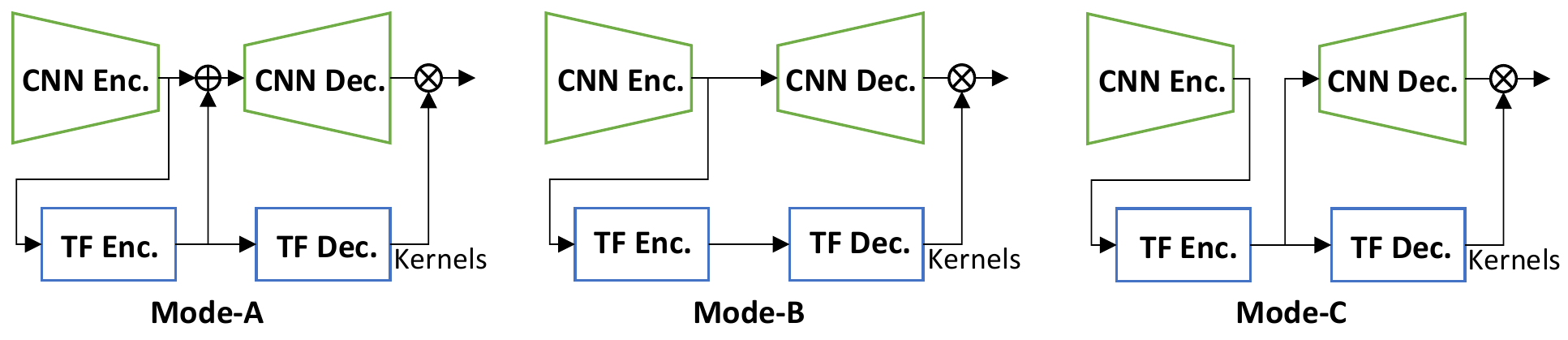}
\end{center}
\caption{Three variants of TransDoDNet.}
\label{fig:memory_type}
\end{figure}

: 
In this section, we design three variants of TransDoDNet that are different from the usage mode of Transformer encoder features, as shown in Fig.~\ref{fig:memory_type}. The mode-A jointly fuses the CNN features and Transformer features as the input of the CNN decoder; the mode-B only transfers the CNN features into the CNN decoder; and the mode-C only transfers the Transformer features into the CNN decoder. The three variants are compared in Table~\ref{tab:abc}. It demonstrates that the complementarity of both features contributes to better performance than any of them. 

\begin{table}[h]
\small 
\caption{Comparison of three variants of TransDoDNet.}
\begin{center}
  \begin{tabular}{c|c|c}
    \hline
    \# Variants & mDice & mHD \\ \hline
    Mode-A & \textbf{73.79} &  \textbf{18.95} \\ \hline 
    Mode-B & 72.48 & 20.87  \\ \hline
    Mode-C & 72.60 & 20.60 \\ \hline
    \end{tabular}
\end{center}
\label{tab:abc}
\end{table}

\begin{figure*}[h]
  \begin{center}
  \includegraphics[width=1.0\linewidth]{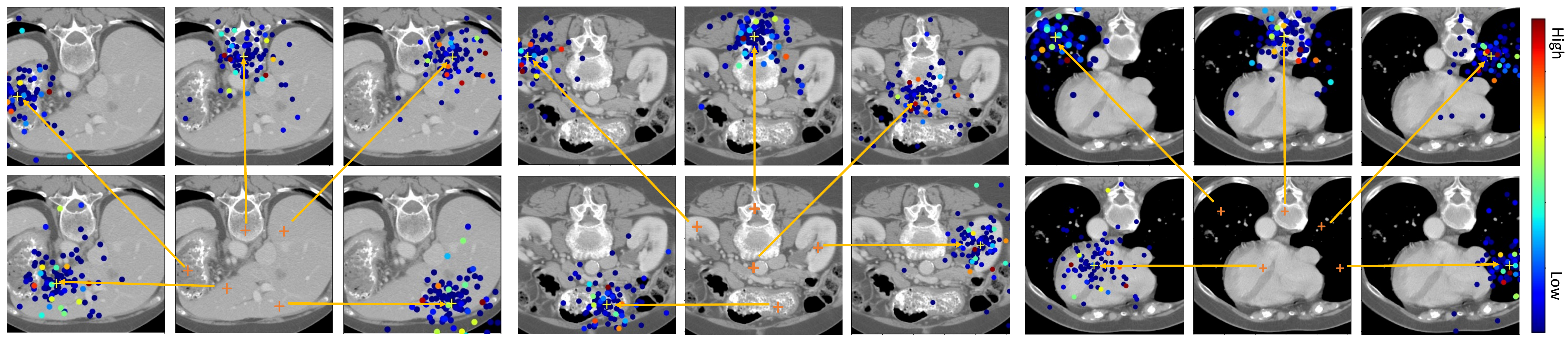}
  \end{center}
  \caption{Deformable self-attention visualization. }
  \label{fig:att_vis}
  \end{figure*}

\begin{table*}
\small
\caption{Performance (Dice, \%, higher is better; HD, lower is better) of different methods on seven partially labeled datasets. Note that `Average score' is the aggregative indicator that averages the Dice or HD over 11 categories.}
\begin{center}
\begin{tabular}
{c|m{0.76cm}<{\centering}|m{0.76cm}<{\centering}|m{0.76cm}<{\centering}|m{0.76cm}<{\centering}|m{0.76cm}<{\centering}|m{0.76cm}<{\centering}|m{0.76cm}<{\centering}|m{0.76cm}<{\centering}|m{0.76cm}<{\centering}|m{0.76cm}<{\centering}|m{0.76cm}<{\centering}|m{0.76cm}<{\centering}}
\hline
\multirow{3}{*}{Methods} & \multicolumn{4}{c|}{Task 1: Liver} &  \multicolumn{4}{c|}{Task 2: Kidney} & \multicolumn{4}{c}{Task 3: Hepatic Vessel} \\ \cline{2-13} & \multicolumn{2}{c|}{Dice} & \multicolumn{2}{c|}{HD} & \multicolumn{2}{c|}{Dice} & \multicolumn{2}{c|}{HD} & \multicolumn{2}{c|}{Dice} & \multicolumn{2}{c}{HD} \\ \cline{2-13} 
  & Organ & Tumor & Organ & Tumor & Organ & Tumor & Organ & Tumor & Organ & Tumor & Organ & Tumor \\ \hline
Multi-Nets & 96.54 & 63.9 & 4.91 & 34.14 & 96.52 & 78.07 & 2.96 & 9.01 & 62.85 & 71.56 & 12.35 & 35.63 \\ \hline
TAL \cite{fang2020multi} & 96.21 & 64.1 & 5.53 & 33.21 & 96.01 & 78.87 & 3.38 & 8.63 & 64.64 & 73.87 & 11.56 & 28.36 \\ \hline
Multi-Head \cite{chen2019med3d} & 96.77 & 64.33 & 4.61 & 30.75 & 96.84 & 82.39 & 2.58 & 6.78 & 65.21 & 74.62 & 11.13 & 26.73 \\ \hline
Cond-Input \cite{chen2017fast} & 96.67 & 65.61 & 4.79 & 28.19 & 96.97 & 83.17 & 2.14 & 6.28 & 65.14 & 74.98 & 11.34 & 24.39 \\ \hline
Cond-Dec \cite{dmitriev2019learning} & 96.23 & 65.25 & 5.28 & 29.63 & 96.43 & 83.43 & 2.98 & 5.97 & 65.38 & 72.26 & 10.98 & 28.79 \\ \hline
DoDNet & 96.86 & 65.99 & 3.88 & 27.95 & 97.31 & 83.45 & 1.68 & 4.47 & 65.80 & 77.07 & 10.51 & 32.96 \\ \hline
TransDoDNet & 97.01 & 66.27 & 3.47 & 24.94 & 97.03 & 83.57 & 1.76 & 4.65 & 65.43 & 76.78 & 10.83 & 22.57 \\ \hline \hline
\multirow{3}{*}{Methods} & \multicolumn{4}{c|}{Task 4: Pancreas}               & \multicolumn{2}{c|}{Task 5: Colon} & \multicolumn{2}{c|}{Task 6: Lung} & \multicolumn{2}{c|}{Task 7: Spleen} & \multicolumn{2}{c}{Average score}              \\ \cline{2-13} 
  & \multicolumn{2}{c|}{Dice} & \multicolumn{2}{c|}{HD} & Dice             & HD              & Dice            & HD              & Dice             & HD               & \multirow{2}{*}{mDice} & \multirow{2}{*}{mHD} \\ \cline{2-11}
  & Organ & Tumor & Organ & Tumor & Tumor & Tumor & Tumor & Tumor & Organ & Organ & & \\ \hline
Multi-Nets & 83.18 & 56.17 & 6.32 & 18.41 & 42.39 & 72.56 & 61.68 & 40.18 & 94.37 & 2.18 & 73.38 & 21.70 \\ \hline
TAL \cite{fang2020multi} & 83.39 & 60.93 & 5.35 & 9.56 & 45.14 & 57.98 & 67.59 & 21.51 & 94.72 & 2.07 & 75.04 & 17.01 \\ \hline
Multi-Head \cite{chen2019med3d} & 84.59 & 64.11 & 4.19 & 8.73 & 46.65 & 40.67 & 68.63 & 20.01 & 95.47 & 1.51 & 76.33 & 14.34 \\ \hline
Cond-Input \cite{chen2017fast} & 84.24 & 63.32 & 5.87 & 8.91 & 46.03 & 42.45 & 69.67 & 19.49 & 95.15 & 1.59 & 76.45 & 14.13 \\ \hline
Cond-Dec \cite{dmitriev2019learning} & 83.86 & 62.97 & 5.22 & 9.12 & 50.67 & 34.54 & 63.77 & 35.15 & 94.33 & 2.78 & 75.87 & 15.49 \\ \hline
DoDNet & 85.39 & 60.22 & 5.51 & 9.31 & 47.66 & 36.39 & 72.65 & 7.12 & 93.94 & 2.92 & 76.94 & 12.97 \\ \hline
TransDoDNet & 84.88 & 64.63 & 6.26 & 8.34 & 58.64 & 27.26 & 70.82 & 14.28 & 95.82 & 1.38 & 78.26 & 11.43 \\ \hline
\end{tabular}
\end{center}
\label{tab:SOTA}
\end{table*}

\subsection{Attention Visualization}
In Fig.~\ref{fig:att_vis}, we visualize the sampling key points of different reference points in the deformable self-attention module of the last Transformer encoder layer. 
For the sake of convenience in visualization, we display the sampling points at multi-scale levels on a CT slice. The sampling points and reference points is refers to the circle dots and cross signs, respectively. 
And the different colors of sampling points denote the different levels of attention weights. 
It shows that the deformable self-attention can adaptively adjust the attention key points to the locations nearby or associated with query objects.

\subsection{Comparing to State-of-the-art Methods}

In this section, we compare the proposed TransDoDNet to the state-of-the-art competitors, which also attempt to address the partially labeling issue, on seven partially labeled tasks using the MOTS test set. 
These methods include (1) seven individual networks, each being trained on a partially dataset (denoted by Multi-Nets), (2) two multi-head networks (\textit{i.e.}, Multi-Head~\cite{chen2019med3d} and TAL~\cite{fang2020multi}), (3) two single-network methods with the task condition (\textit{i.e.}, Cond-Input~\cite{chen2017fast}, Cond-Dec~\cite{dmitriev2019learning}), and DoDNet with convolution as kernel generator~\cite{zhang2021dodnet}. 
To ensure a fair comparison, we keep the same encoder-decoder architecture for all methods, except that the channels of CNN decoder layers in Multi-Head are halved due to the GPU memory limitation. 

Table~\ref{tab:SOTA} shows the performance metrics for the segmentation of each organ$/$tumor and the average scores over 11 categories. It reveals that
(1) almost all the methods (TAL, Multi-Head, Cond-Input, Cond-dec, DoDNet, TransDoDNet) achieve better performance than seven individual networks (Multi-Nets), suggesting that training with more data (even partially labeled) is beneficial to the performance;
(2) the dynamic filter generation strategy is superior to directly embedding the task condition into the input or decoder (used in Cond-Input and Cond-Dec); 
(3) Compared to the DoDNet with the convolution based kernel generator, our TransDoDNet with the Transformer based kernel generator further improves the Dice by 1.32\% and reduces the HD by 1.54; and
(4) the proposed TransDoDNet achieves the highest overall performance with a mDice of 78.26\% and a mHD of 11.43, beating all competitors. 
To make a qualitative comparison, we visualize the segmentation results obtained by seven methods on seven tasks in Figure~\ref{fig:visual}. It shows that our TransDoDNet outperforms other methods, especially in segmenting small tumors. 

\begin{figure}[t]
  \begin{center}
  \includegraphics[width=0.8\linewidth]{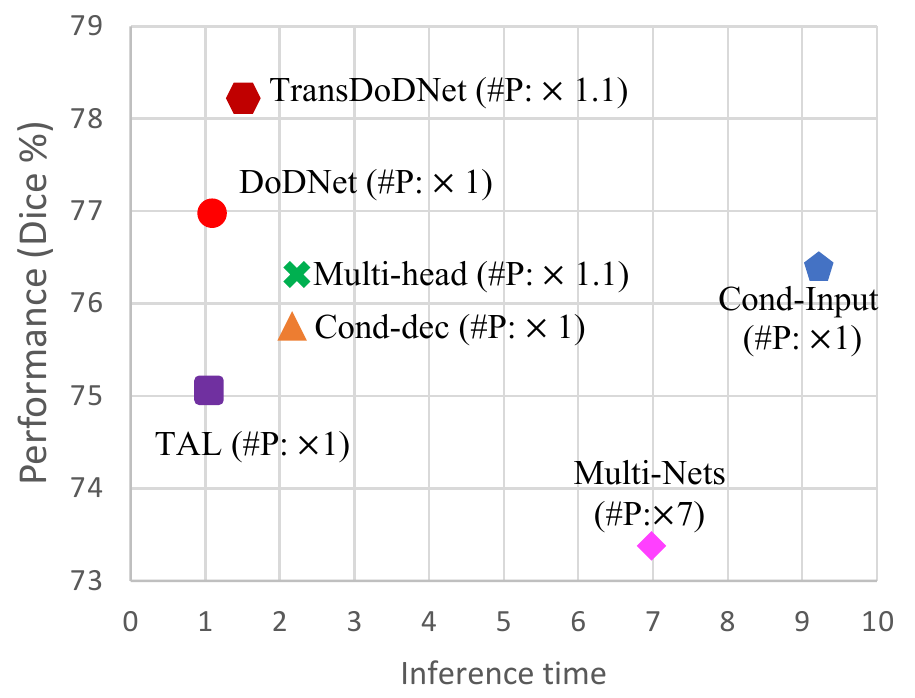}
  \end{center}
  \caption{Accuracy vs.\ complexity. The accuracy refers to the overall Dice score on the MOTS test set. The complexity refers to the inference time cost. Both inference time and model parameters (`\#P') are quantified based on a single network that can only process a single partial segmentation task. 
  }
  \label{fig:inference_time}
  \end{figure}

\begin{figure*}[t]
\begin{center}
\includegraphics[width=1.0\linewidth]{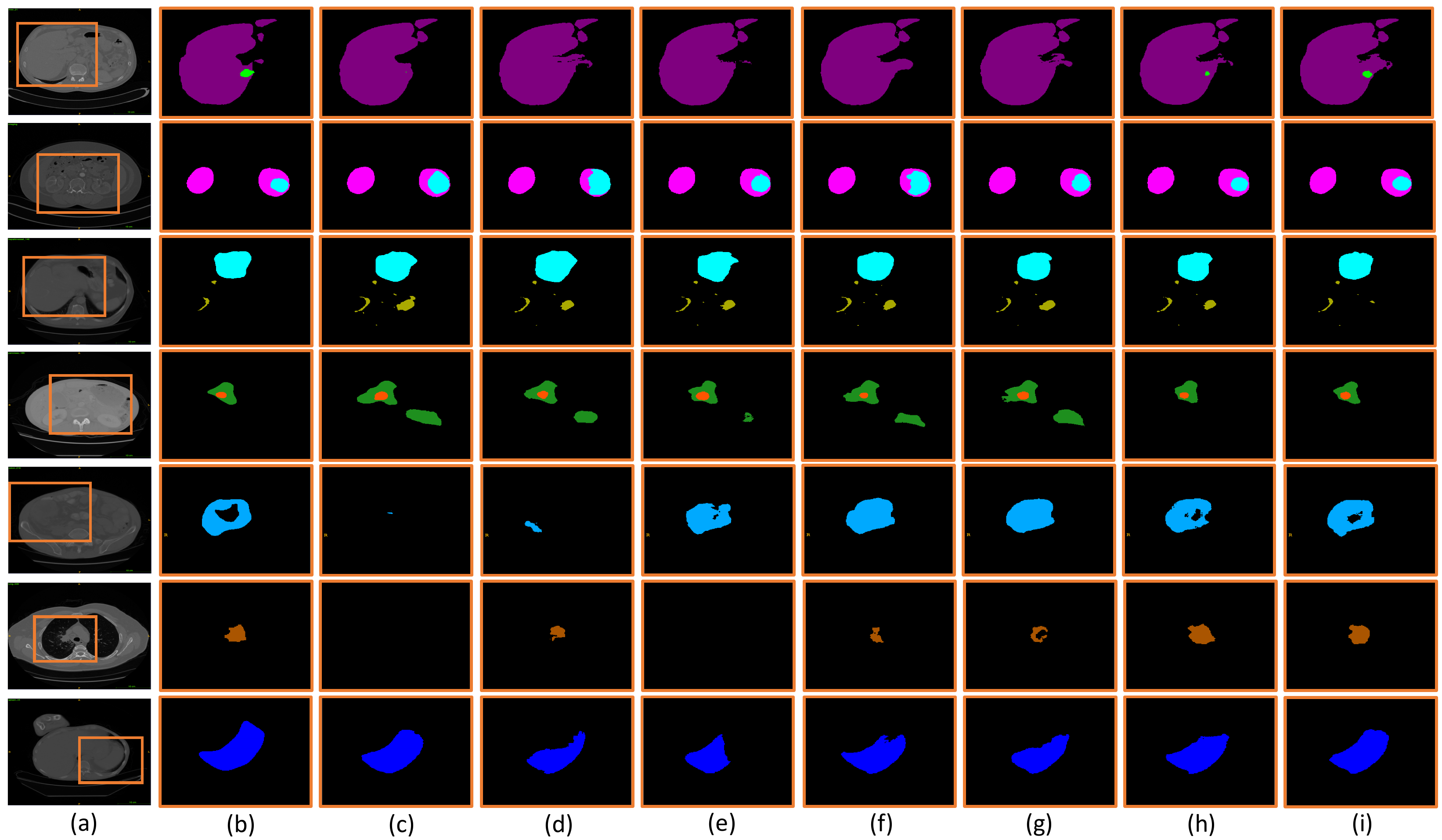}
\end{center}
\caption{Visualization of (a) input image and (b) its ground truth, and segmentation results obtained by (c) Multi-Nets; (d) TAL~\cite{fang2020multi}; (e) Multi-Head~\cite{chen2019med3d}; (f) Cond-Input~\cite{chen2017fast}; (g) Cond-Dec~\cite{dmitriev2019learning}; (h) DoDNet; and (i) TransDoDNet. 
}
\label{fig:visual}
\end{figure*}

\label{Sec.SpeedvsAccuracy} 
In Figure~\ref{fig:inference_time}, we also compare the accuracy-complexity trade-off of seven methods. 
Most of the methods, including TAL, Cond-Dec, Cond-Input, DoDNet, and TransDoDNet, share the encoder and decoder for all tasks, and hence have a similar number of parameters as one single network. Although our TransDoDNet has an extra Transformer module, the number of parameters assigned for this part is negligible compared to the CNN encoder. 
The Multi-Head network has a few more parameters due to the use of multiple task-specific decoders. 
Multi-Nets has to train seven single networks to address these partially labeled tasks individually, resulting in seven times more parameters than a single network.
As for the inference complexity, Cond-Input, Multi-Nets, Multi-Head, and Cond-Dec suffer from the repeated inference processes, and hence need more time to perform seven partial segmentation tasks than other methods. Besides, embedding task information in the input layer leads to the additional increased complexity in the Cond-Input. 
In contrast, TAL is much more efficient in segmenting all targets, since the encoder-decoder (except for the last segmentation layer) is shared by all tasks.
Both DoDNet and TransDoDNet share the encoder-decoder architecture and specialize the dynamic head for each partially labeled task. The inference of the dynamic head is very fast due to its lightweight architecture. 
Although the Transformer module in TransDoDNet results in the extra computation complexity, it achieves the best performance with an acceptable inference complexity. 

\subsection{MOTS Pre-training for Downstream Tasks}

It has been generally recognized that training a deep model with more data contributes to a better generalization ability \cite{zoph2020rethinking,he2019rethinking,tran2018closer,esteva2017dermatologist}.
However, deep learning remains trammeled by the limited annotations, especially in the 3D medical image segmentation. 
Although self-supervised learning \cite{ssl_survey_pami,SSL_survey_arXiv,PIPL,Transformation,CMC,MOCO,SimCLR,BYOL} has shown great potential to address this issue, it still suffers from the lack of a big dataset with millions or trillions of free data for self-supervised learning, especially for 3D volumes. 
With one thousand strong labels, we believe that MOTS is able to produce a more effective pre-trained model for the annotation-limited downstream tasks. 
To demonstrate this, we compare the pre-trained weights on MOTS to other pre-train methods, including (1) SinglePT, pre-training on a single dataset; (2) BYOL, a predominant self-supervised learning method \cite{BYOL}. 
We choose the largest Hepatic Vessel segmentation benchmark as the single pre-trained dataset in SinglePT, considering it contains up to 242 training cases. 
For the self-supervised learning purpose, we collect the unlabeled CT volumes from public datasets as much as possible, resulting in an unlabeled dataset with more than three thousand volumes. 
We extend the BYOL method to a 3D self-supervised learning framework as done in~\cite{xie2021unified} and pre-train it on the unlabeled CT volumes. 

\begin{figure}[t]
\begin{center}
\includegraphics[width=0.8\linewidth]{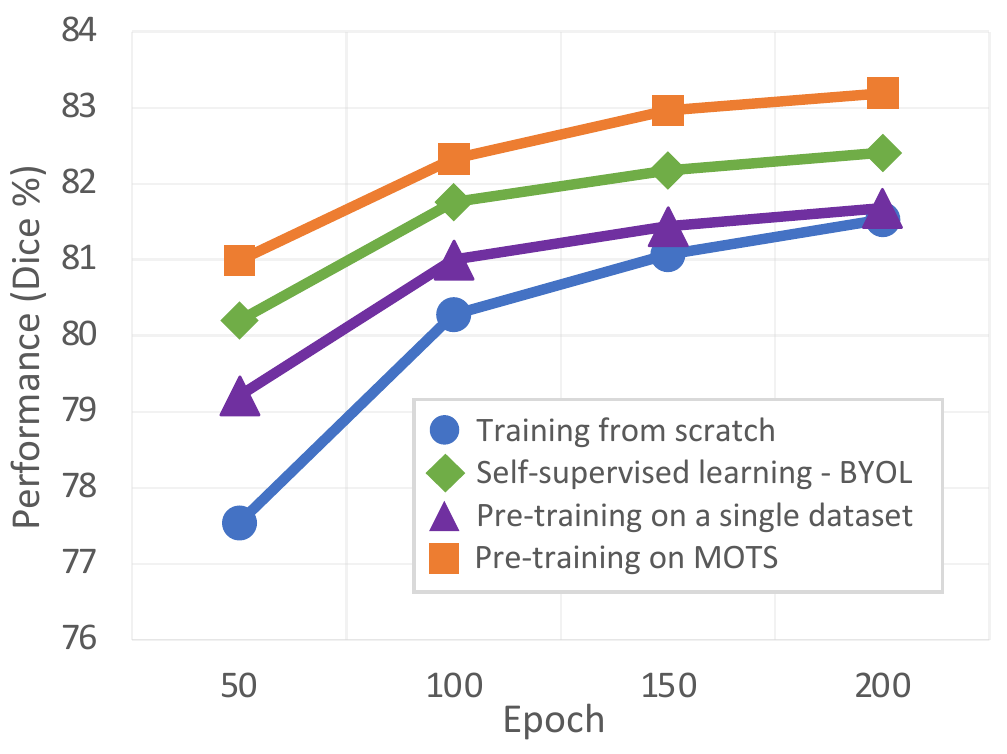}
\end{center}
\caption{Comparison of validation performance of four different initialization strategies, including training from scratch, pre-training on a single dataset \#3 Hepatic Vessel, self-supervised learning (BYOL), and our MOTS partially supervised pre-training. Here the validation performance refers to the averaged Dice score over 13 categories.
}
\label{fig:Downstream_valdice}
\end{figure}

\begin{figure}[h]
\begin{center}
\includegraphics[width=1.0\linewidth]{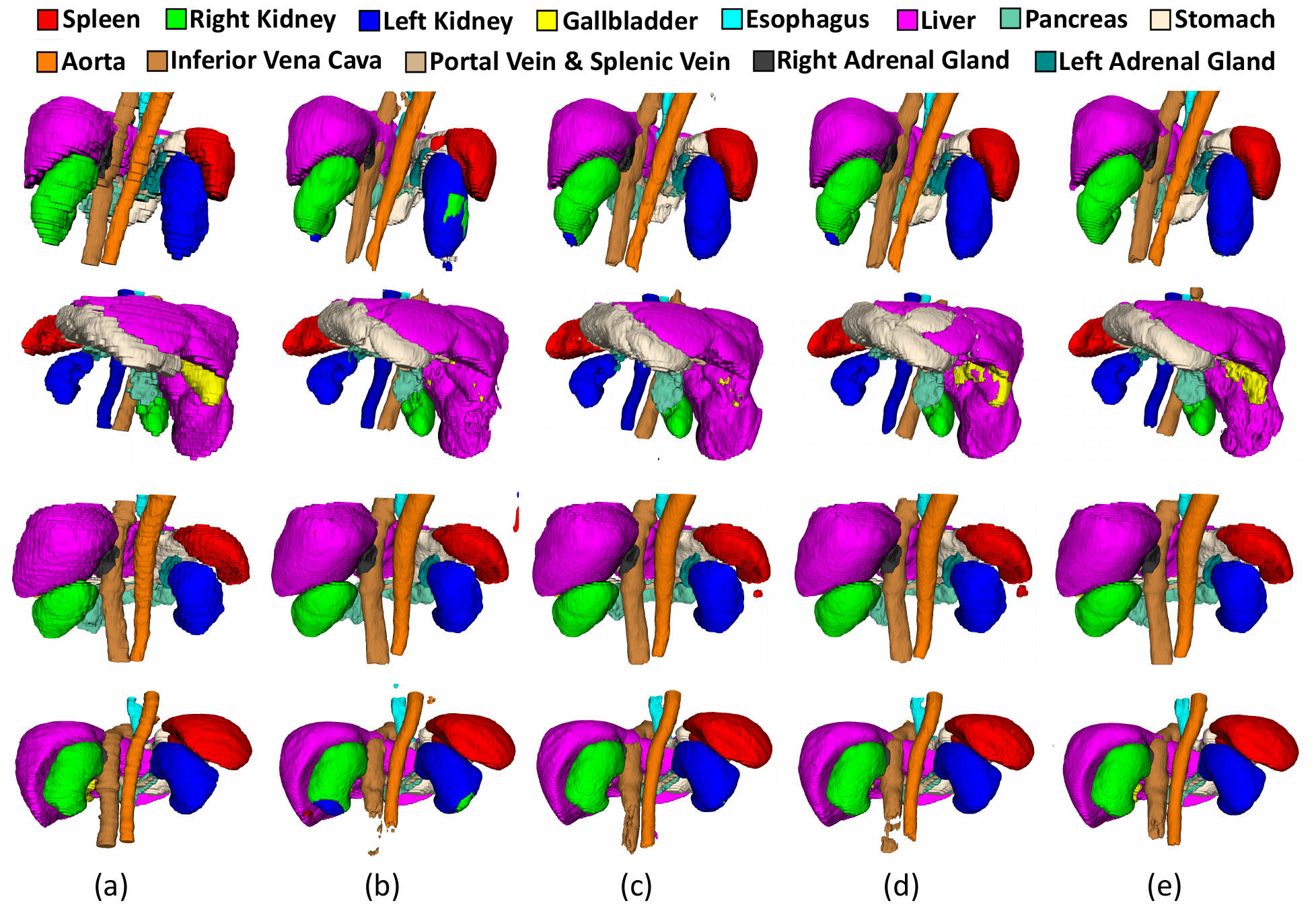}
\end{center}
\caption{Visualization of results obtained by applying the segmentation network with four initialization strategies. (a) Ground truth; (b) random initialization; (c) pre-trained weights on the single dataset \#3 Hepatic Vessel; (d) self-supervised learning on the unlabeled dataset; and (e) pre-trained weights on MOTS.}
\label{fig:BCV_visual}
\end{figure}

\noindent \textbf{Transferring to BCV}: 
We initialize the segmentation network, which has the same encoder-decoder structure as introduced in Sec.~\ref{Sec.encoder-decoder}, using three initialization strategies, including random initialization (\textit{i.e.}, training from scratch), pre-training on SinglePT, and pre-training on MOTS.
We split half of the cases from the BCV training set for validation since the annotations of the BCV test set are withheld for online assessment, which is inconvenient.
We compare the validation performance of four different initialization methods during the training, shown in Fig.~\ref{fig:Downstream_valdice}. 
It reveals that, compared to training from scratch, all pre-training methods help the model converge quickly and perform better, particularly in the initial stage.
However, pre-training on a single dataset (\textit{i.e.}, \#3 Hepatic Vessel) only slightly outperforms training from scratch. 
Although the unlabeled data is twice as much as MOTS, it is still challenging for self-supervised learning to explore strong representation ability without strong supervision. 
Not surprisingly, pre-training on MOTS achieves not only the fastest convergence but also a remarkable performance boost. 
The quantitative results are compared in Table~\ref{tab:BCV_val}. It reveals that pre-training on MOTS has a strong generalization potential that achieves 83.19\% mDice, outperforming the baseline model (TFS) by 1.66\%, outperforming the self-supervised learning BYOL by 0.88\%. 
As for a qualitative comparison, we also visualize the 3D segmentation results obtained by using different initialization strategies in Fig.~\ref{fig:BCV_visual}. It shows that initializing the segmentation network with the MOTS pre-trained weights enables the model to achieve more accurate segmentation results than other initialization. 
Both quantitative and qualitative results demonstrate the strong generalization ability of the model pre-trained on MOTS.


\begin{table}[]
  \caption{Comparison of different pre-trained methods on the BCV validation set. TFS: training network from scratch; SinglePT: pre-training on a single dataset; BYOL: self-supervised learning on the unlabeled dataset; MOTS*: DoDNet pre-trained on MOTS; MOTS: TransDoDNet pre-trained on MOTS.}
  \centering
  \begin{tabular}{c|c|c}
  \hline
  Method   & mDice & mHD   \\ \hline
  TFS      & 81.53 & 8.14 \\ \hline
  SinglePT & 81.68 & 8.02 \\ \hline
  BYOL~\cite{BYOL}     & 82.31 & 7.89  \\ \hline
  MOTS*    & 82.51  & 7.77  \\ \hline
  MOTS     & \textbf{83.19} & \textbf{7.44}  \\ \hline
  \end{tabular}
  \label{tab:BCV_val}
  \end{table}

Furthermore, we also evaluate the effectiveness of the MOTS pre-trained weights on the BCV unseen test set. 
We compare our method to other state-of-the-art methods in Table~\ref{tab:BCV_test}, including Auto Context~\cite{roth2018multi}, deedsJointCL~\cite{deedsJointCL}, DLTK~\cite{pawlowski2017dltk}, PaNN~\cite{zhou2019prior}, and nnUnet~\cite{isensee2021nnu}. 
Compared to the training from scratch, using the MOTS pre-trained weights contributes to a substantial performance gain, improving the average Dice from 87.28\% to 88.21\%, reducing the average HD from 14.55 to 13.11, and reducing the average mean surface distance (SD) from 1.49 to 0.89. 
After making the TransDoDNet as the pre-training model, we further boost the performance of all metrics, setting state-of-the-art performance on the test set, \textit{i.e.}, mDice of 88.60\%, mHD of 12.83, and mSD of 0.86. 

\begin{table}
\caption{Comparison of state-of-the-art methods on the BCV test set. SD: Mean surface distance (lower is better); TFS: Training network from scratch; MOTS*: DoDNet pre-trained on MOTS; MOTS: TransDoDNet pre-trained on MOTS. The values of three metrics are averaged over 13 categories.}
\begin{center}
\small 
\begin{tabular}{c|c|c|c}
\hline
\small 
Methods & mDice & mHD & mSD \\ \hline
Auto Context \cite{roth2018multi} & 78.24 & 26.10 & 1.94 \\ \hline
deedsJointCL \cite{deedsJointCL} & 79.00 & 25.50 & 2.26 \\ \hline
DLTK \cite{pawlowski2017dltk} & 81.54 & 62.87 & 1.86 \\ \hline
PaNN \cite{zhou2019prior} & 84.97 & 18.47 & 1.45 \\ \hline
nnUnet \cite{isensee2021nnu} & 88.10 & 17.26 & 1.39 \\ \hline
TFS & 87.28  & 14.55 & 1.49 \\ \hline
MOTS*  & 88.21 & 13.11 & 0.89 \\ \hline
MOTS  & \textbf{88.60} & \textbf{12.83} & \textbf{0.86} \\ \hline
\end{tabular}
\end{center}
\label{tab:BCV_test}
\end{table}

\noindent \textbf{Transferring to other modalities}: 
We also attempt a more challenging task, \textit{i.e.}, transferring MOTS pre-trained weights to BraTS that has totally different modalities.
Considering the brain MRI with four channels, we extend the channels in the first layer and initialize them by copying the weights of the first channel. 
We compare the segmentation performance of MOTS pre-trained model with the nine competitive methods, including CascadeUNet \cite{wang2018automatic}, DMFNet \cite{chen2019dmfnet}, OM-Net \cite{zhou2020oneTIP}, DeepSCAN \cite{mckinley2018ensembles}, VAE-Seg\cite{myronenko20183d}, ReversibleUnet \cite{ReversibleUnet}, DCAN \cite{DCAN}, nnUnet \cite{isensee2021nnu}, and ConResNet \cite{zhang2020conresnet}. 
Table~\ref{tab:brats} shows the segmentation of different models on the BraTS online evaluation set. Note that all segmentation results are evaluated online, and the performance of competing models is adopted from the literature.
It reveals that compared to the training network from the scratch method, our MOTS pre-trained model improves the average Dice from 85.74\% to 86.18\% and reduces the average HD from 4.74 to 4.39, resulting in a competitive performance. 
More importantly, such a competitive performance is obtained by using a single model, instead of the ensembles used by most of the competitors. 

\begin{table*}[t]
\centering
\caption{Comparison to the state-of-the-art segmentation methods on the BraTS dataset.}
\begin{tabular}{c|c|m{0.76cm}<{\centering}|m{0.76cm}<{\centering}|m{0.76cm}<{\centering}|m{0.76cm}<{\centering}|m{0.76cm}<{\centering}|m{0.76cm}<{\centering}|m{0.76cm}<{\centering}|m{0.76cm}<{\centering}}
\hline
\multirow{2}{*}{Methods}  & \multirow{2}{*}{Ensembles} & \multicolumn{3}{c|}{Dice score (\%)} & \multicolumn{3}{c|}{Hausdorff distance (HD)} & \multirow{2}{*}{mDice} & \multirow{2}{*}{mHD} \\ \cline{3-8}
& & ET & WT & TC & ET & WT & TC & &  \\ \hline
CascadeUNet \cite{wang2018automatic} & $\surd$ & 79.72 & 90.21 & 85.83 & 3.13 & 6.18 & 6.37 & 85.25 & 5.23 \\ \hline
DMFNet \cite{chen2019dmfnet} & $\surd$ & 80.12 & 90.62 & 84.54 & 3.06 & 4.66 & 6.44 & 85.09 & 4.72 \\ \hline
OM-Net \cite{zhou2020oneTIP} & $\surd$ & 81.11 & 90.78  & 85.75 & 2.88 & 4.88 & 6.93 & 85.88 & 4.90 \\ \hline
DeepSCAN \cite{mckinley2018ensembles} & $\surd$ & 79.60 & 90.30 & 84.70 & 3.55 & 4.17 & 4.93 & 84.87 & 4.22\\ \hline
VAE-Seg \cite{myronenko20183d} & $\surd$ & 82.33 & 91.00 & 86.68 & 3.93 & 4.52 & 6.85 & 86.67 & 5.10 \\ \hline
DCAN \cite{DCAN} & $\surd$ & 81.71 & 91.18 & 86.19 & 3.57 & 4.26 & 6.13 & 86.36 & 4.65 \\ \hline
nnUnet \cite{isensee2021nnu} & $\surd$ & 80.87 & 91.26 & 86.34 & 2.41 & 4.27 & 6.52 & 86.16 & 4.40 \\ \hline
ConResNet \cite{zhang2020conresnet} & $\surd$ & 83.15 & 90.38 & 85.90 & 2.79 & 4.15 & 5.75 &  86.48 & 4.23 \\ \hline \hline
ConResNet \cite{zhang2020conresnet} & $\times$ & 81.37 & 90.22 & 85.28 & 3.09 & 4.55 & 6.08 &  85.62 & 4.57 \\ \hline
ReversibleUnet \cite{ReversibleUnet}  & $\times$ & 80.33   & 91.01  & 86.21   & 2.58 & 4.58 & 6.84 & 85.85 & 4.67 \\ \hline
TFS & $\times$ & 80.33 & 90.54 & 86.36 & 3.34 & 4.78 & 6.10 & 85.74 & 4.74 \\ \hline
MOTS & $\times$ & 81.10 & 90.93 & 86.52 & 3.01 & 4.23 & 5.93 & 86.18 & 4.39 \\ \hline
\end{tabular}
\label{tab:brats}
\end{table*}

\section{Conclusion}
In this paper, we create a large-scale multi-organ and tumor segmentation benchmark (MOTS) that integrates multiple partially labeled medical image segmentation datasets. 
To address the partially labeling issue of MOTS, we propose the TransDoDNet model that employs the dynamic segmentation head to flexibly process multiple segmentation tasks. 
Moreover, we introduce a Transformer based kernel generator that models the organ-wise dependencies by using the self-attention mechanism. 
Our experiment results on MOTS indicate that the TransDoDNet model achieves the best overall performance on seven organ and tumor segmentation tasks. 
We also demonstrate the value of the TransDoDNet and MOTS dataset by successfully transferring the weights pre-trained on MOTS to downstream tasks that only limited annotations are available. It suggests that the byproduct of this work (\textit{i.e.}, a pre-trained 3D network, including both encoder and decoder) is conducive to other small-sample 3D medical image segmentation tasks.

\ifCLASSOPTIONcaptionsoff
  \newpage
\fi

\bibliographystyle{IEEEtran}
\bibliography{egbib}

\end{document}